# Semi Automatic Color Segmentation of Document Pages

Application to administrative document segmentation

Stéphane Bres, Véronique Eglin
LIRIS UMR CNRS 5205- INSA Lyon
69621 Villeurbanne Cedex - FRANCE
{stephane.bres, veronique.eglin}@liris.cnrs.fr

Vincent Poulain d'Andecy
ITESOFT
30470 Aimargues - FRANCE
Vincent.PoulaindAndecy@itesoft.com

*Abstract*—This paper presents a semi automatic method used to segment color documents into different uniform color plans. The practical application is dedicated to administrative documents segmentation. In these documents, like in many other cases, color has a semantic meaning: it is then possible to identify some specific regions like manual annotations, rubber stamps or colored highlighting. A first step of user-controlled learning of the desired color plans is made on few sample documents. An automatic process can then be performed on the much bigger set as a batch. Our experiments show very interesting results in with a very competitive processing time.

## I. INTRODUCTION

The work we present here was developed during the project PiXL[1], in collaboration with the ITSOFT society, which is specialized in automatic document analysis. The aim of this work is to detect and process all the information carried by the color and the color variations, in order to allow the separation of the different physical structured regions.

### A. Color information and semantic

Scanning color documents is strongly recommended whenever it is possible. A color image contains about 65,000 times more information than an image in gray scale, with a file weight not exceeding 3 times the gray scale version (raw data). The color in documents is used both for structural and decorative purposes. Documents rendering and legibility of information is generally better. In order to bring out the different available information in the document, the colorimetric analysis of color images is an interesting and robust approach. Indeed, with the assumption, widely verified that a homogeneous area or overall color plot corresponds to a semantic unit, a pixels based classification that relies on the association of colorimetric and spatial information of pixels is an interesting way to obtain a satisfying separation into different semantic colored area elements (like rubber stamps, handwritten notes, various markings, glued stickers, highlighting,…). Document segmentation into homogeneous regions in terms of color (and thus content) is a major challenge in document analysis. This study is based on test documents provided by ITESOFT. They are representative of a wide variety of difficulties on which we have paid particular attention.

### B. Main causes of segmentation problems

The printing methods are not standardized: different sampling levels can be used resulting in visual distortions that have a degrading effect on the color rendering of the document. As a matter of fact, printers often simulate the original colors using the CMYK color model. The human eye does not perceive this change very well, though the scanner is sensitive and often detects it. Thus, according to the width of the framing and the scan resolution, very annoying artifacts may appear. Other distortions, noise and false colors are due to the process of scanning and JPEG compression. It reduces color consistency layers. More specifically, JPEG compression is the cause of mixture between text and background colors, leading to an over-segmentation of the color classes. The scanning process impacts the contrast between text and background regions. The quality of paper and the various degradations due to too intensive use of documents may also complicate the analysis. The initial framing of the document and all regions with linear variations of color (presence of continuous degraded in some bands) can lead to false detections color classes for the same region. The presence of graphic, curves and color textures (sometimes implicitly or in composite backgrounds) within the document leads to less accurate analysis. Furthermore, for any industrialization, the large size of the scanned documents (about 2500 x 3500 pixels) must be taken into account when choosing segmentation techniques. Consequently, fast treatments are needed. This constraint reduces the choice of possible approaches and in a certain way the quality we can expect for the color segmentation process.

## II. OUR SEMI-AUTOMATIC APPROACH

There are possible clustering methods to form $k$ sub-spaces corresponding to $k$ color classes from the image color space according to a similarity criterion pixel colors. These classes are presented as scatter plots of colors associated with the pixels of the image. We can group them according to two families of clustering and classification approaches: with and without learning. For the first category, we can cite the Bayes method [1] Vectors Supports machines (SVM [2] and LS-SVM

[1] PIXL (2013-2016) : french projet funded by the *Programme des Investissements d'avenir*

[3] using the color information and texture), a particular case of Neural Networks Multi Layer [4] (MLP, SOM, etc.), fuzzy Kohonen [5], the k nearest neighbor (k-PPV), etc. Also in the area of the document neuronal approaches [6] and approaches using graphs [7] have been developed to reduce the number of colors in order to extract the text in color. Smigiel [8] proposes a segmentation into four classes (background, text, colored text and text verso) by filing via Kohonen maps. The neural network is trained on a portion of a page to be subsequently run on a whole book.

Concerning the classification without learning, we will quote classifiers based on the method of Fisher [9], the k-means [10,11] WK-means (with selection of weight) [12], the fuzzy C-means [13] (unsupervised), which adapt to the quality of the printing (a comparative study by Mingoti-2006 shows that this technique is more efficient than those of K-means and neural networks [14]), There is a strong tendency these last years the use of a MeanShift classification [15,16] but it is a too time consuming approach for us.

In our approach, we focused on the most accurate and fast possible classification methods. Full automatic methods were tested but with too slow responses or not satisfying or both. We tried then a semi automatic approach which is based on two different steps :

A first step of learning in which the user train a K-means classifier by selecting windows of free size on the document and indicating the number of classes it contains. Most of these windows contain between 2 and 5 classes, not more.

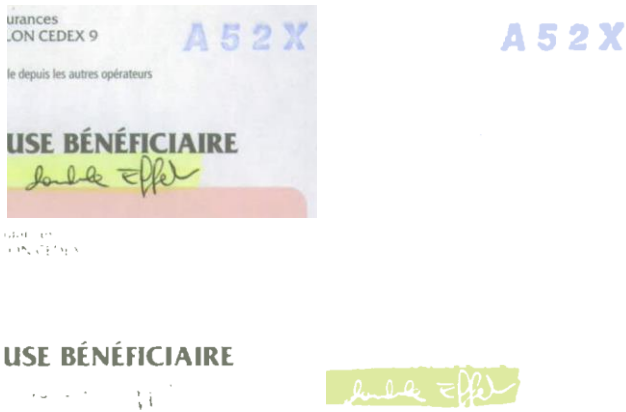

Figure 1. Example of a part of a document and the some of the color planes we obtain after clustering. a) Original image b) rubber stamp c) Printed characters d) Highlighting.

Within less than 5 or 6 windows, it is most of the time possible to define all the color classes of a document. Then a batch treatment is use to apply theses learning classes on many other documents of the same type. An example of the kind of results we obtain is presented on figure 1.

If a new color appears on the document that was not present during the learning process, a group of pixel that are too far from the known color classes is detected and the document is extracted from the automatic process to be treated separately and improve the color description by adding this new color class.

III. EVALUATIONS AND CONCLUSION

The evaluation of this method of segmentation is not that easy. Indeed, we do not have a usable ground truth which is able to quantify the quality of our results. Even if we had a possible segmentation at pixel level, it is not possible to decide that a different proposition (with different classification for some pixels) is worse just because it is different. Consequently, we decide to make a manual quantification of the quality, based on the visual evaluation by different operators. The mean results we obtain are presented below:

Moreover, the different tests we made prove that this approach give very interesting results for the different segmented color planes. For instance, ITESOFT enterprise considers that these results are one of the most interesting in an industrial context. The training process is rather simple and not very time consuming. It can be saved once for all in case of sets of documents of the same type.

ACKNOWLEDGMENT

We would like to thank ITESOFT for their collaboration in the development of this approach. It leads to very rich exchanges for the global design of this segmentation method.